\documentclass[sigconf]{acmart}
\usepackage{tabu}
\usepackage{tabularx}
\usepackage{amsmath}
\usepackage{bbm}
\usepackage{graphicx}
\usepackage{caption}
\usepackage{subcaption}
\usepackage{hyperref}
\usepackage{url}
\usepackage{cases}
\usepackage[english]{babel}
\usepackage{blindtext}
\usepackage{bm}
\usepackage{multirow}
\usepackage{makecell}
\usepackage{balance}

\usepackage{algorithm}
\usepackage{algpseudocode}

\newenvironment{packed_itemize}{
\begin{list}{\labelitemi}{\leftmargin=1.5em}
  \setlength{\itemsep}{1pt}
  \setlength{\parskip}{0pt}
  \setlength{\parsep}{0pt}
  \setlength{\headsep}{0pt}
  \setlength{\topskip}{0pt}
  \setlength{\topmargin}{0pt}
  \setlength{\topsep}{0pt}
  \setlength{\partopsep}{0pt}
}{\end{list}}


\AtBeginDocument{%
 }
\begin{document}

\copyrightyear{2024}
\acmYear{2024}
\setcopyright{acmlicensed}
\acmConference[KDD '24]{Proceedings of the 30th ACM SIGKDD Conference on Knowledge Discovery and Data Mining}{August 25--29, 2024}{Barcelona, Spain}
\acmBooktitle{Proceedings of the 30th ACM SIGKDD Conference on Knowledge Discovery and Data Mining (KDD '24), August 25--29, 2024, Barcelona, Spain}
\acmDOI{10.1145/3637528.3671984}
\acmISBN{979-8-4007-0490-1/24/08}

\title{A Population-to-individual Tuning Framework for Adapting Pretrained LM to On-device User Intent Prediction }

\author{Jiahui Gong}
\orcid{0009-0005-1154-8877}
\affiliation{%
  \institution{Department of Electronic Engineering, BNRist, Tsinghua University}
  \country{Beijing, China}}
\email{gjh22@mails.tsinghua.edu.cn}

\author{Jingtao~Ding$^{*}$}
\thanks{$^{*}$ Corresponding author.}
\orcid{0000-0001-7985-6263}
\affiliation{%
  \institution{Department of Electronic Engineering, BNRist, Tsinghua University}
  \country{Beijing, China}}
\email{dingjt15@tsinghua.org.cn}

\author{Fanjin Meng}
\orcid{0009-0004-5019-4437}
\affiliation{%
  \institution{Department of Electronic Engineering, BNRist, Tsinghua University}
  \country{Beijing, China}}
\email{mengfj23@mails.tsinghua.edu.cn}

\author{Guilong Chen}
\orcid{0009-0000-3749-8912}
\affiliation{%
  \institution{ Honor Device Co., Ltd.}
  \country{Shenzhen, China}}
\email{chenguilong@hihonor.com}

\author{Hong Chen}
\orcid{0009-0009-8279-1443}
\affiliation{%
  \institution{ Honor Device Co., Ltd.}
  \country{Shenzhen, China}}
\email{chenhong3@hihonor.com}

\author{Shen Zhao}
\orcid{0009-0009-1362-0755}
\affiliation{%
  \institution{ Honor Device Co., Ltd.}
  \country{Shenzhen, China}}
\email{zhaoshen@hihonor.com}

\author{Haisheng Lu}
\orcid{0009-0005-8091-1190}
\affiliation{%
  \institution{ Honor Device Co., Ltd.}
  \country{Shenzhen, China}}
\email{luhaisheng@hihonor.com}

\author{Yong Li}
\orcid{0000-0001-5617-1659}
\affiliation{%
  \institution{Department of Electronic Engineering, BNRist, Tsinghua University}
  \country{Beijing, China}}
\email{liyong07@tsinghua.edu.cn}



\renewcommand{\shortauthors}{Jiahui Gong et al.}

\begin{abstract}
Mobile devices, especially smartphones, can support rich functions and have developed into indispensable tools in daily life. With the rise of generative AI services, smartphones can potentially transform into personalized assistants, anticipating user needs and scheduling services accordingly. Predicting user intents on smartphones, and reflecting anticipated activities based on past interactions and context, remains a pivotal step towards this vision. Existing research predominantly focuses on specific domains, neglecting the challenge of modeling diverse event sequences across dynamic contexts. Leveraging pre-trained language models (PLMs) offers a promising avenue, yet adapting PLMs to on-device user intent prediction presents significant challenges. To address these challenges, we propose PITuning, a Population-to-Individual Tuning framework. PITuning enhances common pattern extraction through dynamic event-to-intent transition modeling and addresses long-tailed preferences via adaptive unlearning strategies. Experimental results on real-world datasets demonstrate PITuning's superior intent prediction performance, highlighting its ability to capture long-tailed preferences and its practicality for on-device prediction scenarios.
\end{abstract}


\maketitle
\begingroup\renewcommand\thefootnote{\textsection}

\section{Introduction}




Nowadays mobile devices, especially smartphones, have become a major object that individuals interact with in their daily lives. For example, users use their phones to monitor sleep, wake themselves up, hail a car for commuting, watch short videos in rest time, pay money at restaurants, etc., across most activities in one day. Empowered by the recent booming of generative artificial intelligence~(AI) services~(e.g., chatGPT~\cite{brown2020instructtuning}), the smartphone can further evolve into a personalized assistant that can perceive user needs in advance and timely schedule corresponding services. 
The key pathway toward this future is the capability to predict smartphone users' intents, which refers to what activity they intend to do, based on their previous action sequences and contextual information~\cite{li2022smartphone,li2021understanding}.

Existing works mostly focus on predicting user intents within one specific domain, for example, purchase intent in online platforms~\cite{LiKDD22,ping21kdd,WangIjcai19,LiSIGIR23}, search intent in search engines~\cite{TangWSDM18,yang20iart}, pedestrian intention for robots or autonomous vehicles~\cite{ahmed2019pedestrian,sharma2022pedestrian}. To characterize complex dependencies between intent and context, they leverage specific network architectures including feature interaction networks~\cite{Pu19transformer,Wang2021transformer} or graph neural networks~\cite{WangIjcai19,LiSIGIR23}. In contrast, predicting users' daily activity intent when using smartphones requires modeling diverse event sequences across dynamic changing contexts, which generally rely on large-scale behavioral data. However, with increasing concerns about data privacy leakage and real-time serving latency, real-world prediction applications usually adopt on-device model training and deployment, which adds constraints on data scales and exacerbates the data lacking issue.

Pretrained language models~(PLMs)~\cite{GPT2,brown2020instructtuning}, on the other hand, provide a promising solution owing to their encoded knowledge and commonsense reasoning capability acquired through extensive training on diverse datasets. For example, if someone talks about going jogging every morning, a PLM can infer that the individual values fitness, which might predict other health-related behaviors. In this regard, PLMs have been successfully adapted to other cross-domain tasks related to human behaviors, like recommender systems~\cite{Bao23Tallrec,Geng22P5,wei2023llmrec} and mobility trajectories~\cite{liu2023unitime,jin2023timellm, zhou2023onefitsall}. Therefore, we propose to leverage PLMs for on-device user intent prediction, i.e., adapting P    LMs from the language domain into the daily human behavior domain, which is non-trivial due to the following three challenges:


\begin{itemize}
    \item \textbf{Population-level common behavioral patterns are hard to extract from the noisy aggregation of diverse event sequences.} Predicting user intent based on their previous action events requires the characterization of common transition patterns from event sequences to specific intent. However, not all events correlate to the generation of intent, i.e., information redundancy, and this changes with intent type.
    Although transformer-based architecture has proven its usefulness in the sequential modeling of user events~\cite{sun2019bert4rec, savcisens2023using}, it remains questionable whether common event-intent transition patterns shared among the population can be extracted from the above noisy and redundant event sequences.

    \item \textbf{Individual-level long-tailed preferences are hard to capture by large LMs.} Besides common behavioral patterns, individual preference also matters a lot in predicting user intent. For example, compared with public transport, car-hailing might be a long-tailed choice globally, while favored by a few users. However, long-tailed individual preferences are prone to be overtaken by population-level patterns that dominate the population's behavioral data. This inevitably leads to a biased model favoring those intents with a high proportion after tuning a PLM. Existing works on aligning LM for behavioral modeling~\cite{Shi2023LSAT}  tend to construct tuning tasks analogous to their counterparts in NLP like prompt tuning~\cite{Geng22P5} or instruction tuning~\cite{Bao23Tallrec,Zhang2023Insprom}. Without a specific design, however, it is generally difficult to alleviate the above bias problem of the long-tailed preferences given rather limited individual behavioral data.

    \item \textbf{Designing a practical LM tuning framework to support on-device learning and inference of user intent is difficult.} Existing works have proposed a few cloud-device collaboration approaches to achieve on-device prediction or recommendation, mainly targeting device-side personalization.
    Differently, the expected tuning framework is tailored for pretrained LMs and should be able to leverage large-scale population-level data efficiently and limited individual-level data effectively.
    
\end{itemize}

In this paper, we propose a novel \textbf{P}opulation-to-\textbf{I}ndividual Tuning framework~(named PITuning) for adapting pretrained LM to on-device user intent prediction. The core of the PITuning framework is the population-level behavioral data tuning on a pretrained LM that produces a powerful but gigantic global predictor at the cloud side, and the individual-level tuning that adaptively distills this predictor into a lightweight user-specific predictor at the device side.
To solve the first challenge of extracting common behavioral patterns at the population level, PITuning is designed to better capture dynamic event-to-intent transition patterns, i.e., event-wise information enhancement by an auxiliary event-reconstruction loss and intent-wise attentive modeling on top of pretrained transformer.
As for the second challenge of capturing long-tailed preference distribution at the individual level, PITuning is equipped with a novel unlearning strategy for each user that first identifies a set of intents, which are under-represented in population data but emphasizes unique preferences of this user and then remove the model's memorization on these intents. This further guarantees effectively capturing long-tailed preference by tuning on individual behavioral data at the device side.
To summarize, our main contributions are as follows.

\begin{itemize}
    \item We provide a novel angle of adapting PLMs into the human behavioral domain and further resolve the longstanding issue of capturing long-tailed user preferences.
    
    \item We design a population-to-individual tuning framework for PLM that extracts common behavioral patterns and captures individual unique patterns simultaneously, compatible with on-device prediction scenarios.
    
    \item Experiment results on two real-world datasets demonstrate the superiority of our PITuning over state-of-the-art baselines in terms of intent prediction performance. Notably, the outperformance regarding the macroscopic average of precision and recall is 24\%-37\%, underscoring its capability of capturing long-tailed preference for individuals. Ablation studies and in-depth analysis further support the rationality behind specific method design, as well as the high practicality in terms of efficiency and scalability.
\end{itemize}


\section{Preliminary}
\subsection{Data Analysis}

We begin with a comprehensive data analysis. Initially, we randomly sample 1,000 users to calculate their intent distribution. Subsequently, we employ the KMeans method \cite{kmeans} to cluster users' intent distribution and visualize the result using t-SNE~\cite{tsne}, as illustrated in Figure \ref{fig:tsne}. Additionally, we present the population-level intent distribution alongside the distribution for each cluster. From the figure, we observe that intent distribution varies significantly between clusters. This discrepancy undoubtedly complicates the task of user-personalized modeling.

\begin{figure}[tb]
\vspace{-3mm}
    \centering
    \includegraphics[width = 0.92\linewidth]{ 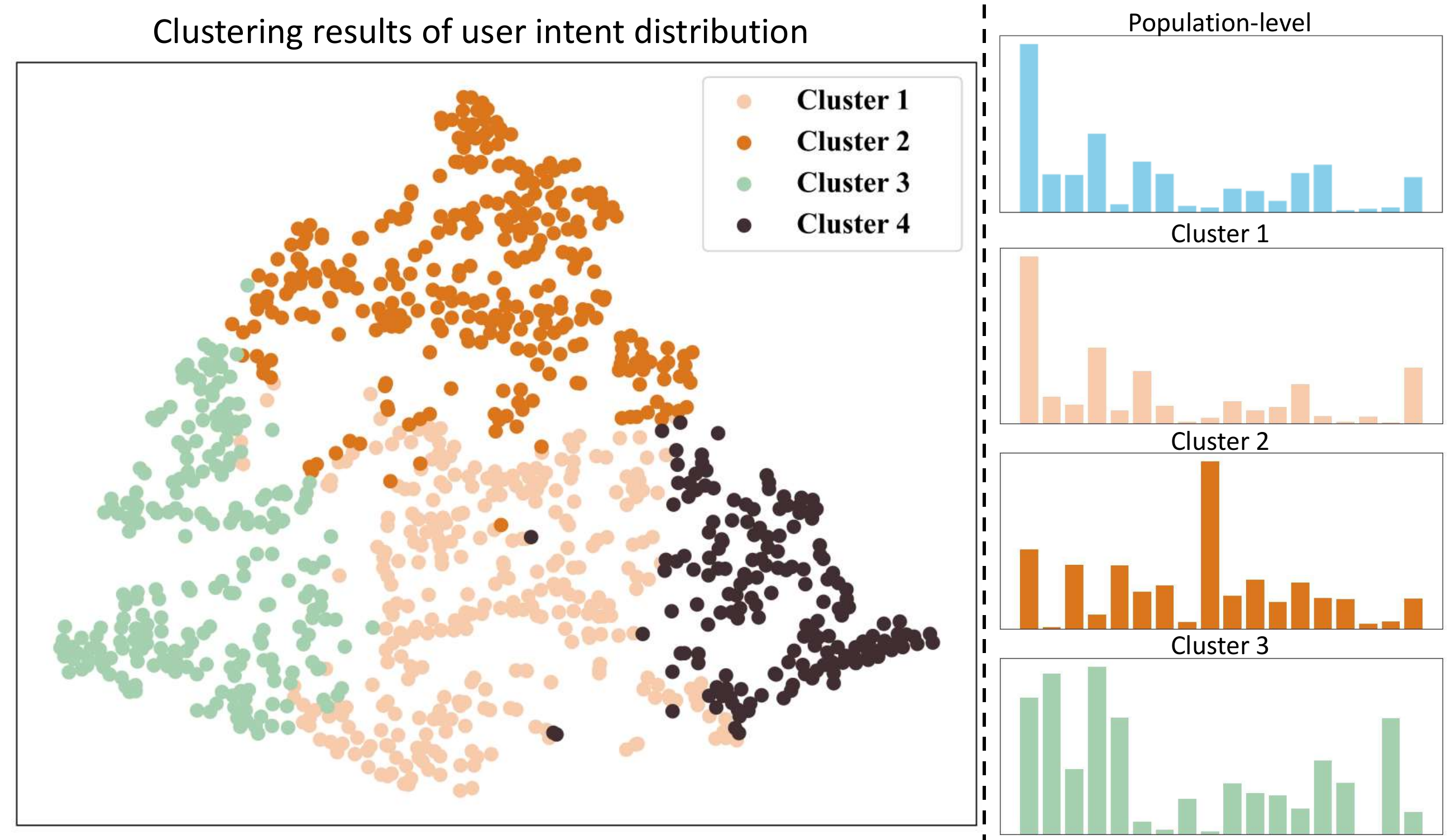}
    \vspace{-3mm}
    \caption{Distribution gap exists between population-level preference and individual-level preference~(by comparing frequency histogram of user intent).}
    \label{fig:tsne}
    \end{figure} 

\subsection{Problem Statement}
Now we give a formal definition of our research problem:

\textsc{Problem1} (User intent prediction). The behavior corresponding to the $i$-th user intention can be represented as $x_i = \left( u_i, l_i, t_i, e_i \right)$, indicating that a specific event $e_i$ takes place involving user $u_i$ at location $l_i$ during time slot $t_i$. Here, $u_i$, $l_i$, $t_i$, and $e_i$ refer to the user ID, location ID, time slot ID, and event ID, respectively. We use $\mathcal{U, L, T, E}$ to denote the sets of users, locations, time slots, and events, with their respective sizes given by $N_U, N_L, N_T,$ and $N_E$. As outlined in the introduction, each user exhibits a particular intention $y_i$ associated with an event-related behavior $x_i$. We define $\mathcal{I}$ as the set of possible intentions, with its size represented by $N_I$.
The event encompasses specific instances involving users, such as the use of app services, spatial trajectory occurrences, and system-related events. The intent captures the underlying goal, purpose, or objective driving these events, effectively grouping them into categories. Therefore, the quantity of distinct intents, denoted by $N_I$, is typically less than the total count of events, represented by $N_E$.

User intent prediction aims to forecast future user intent based on its past $I$ event series, which can be formed as,
\begin{equation}
y_{t} = f(x_{t-I}, x_{t-I+1}, ..., x_{t-1})    
\end{equation}

\begin{figure}[t]
\vspace{-1em}
    \centering
    \includegraphics[width = 0.92\linewidth]{ 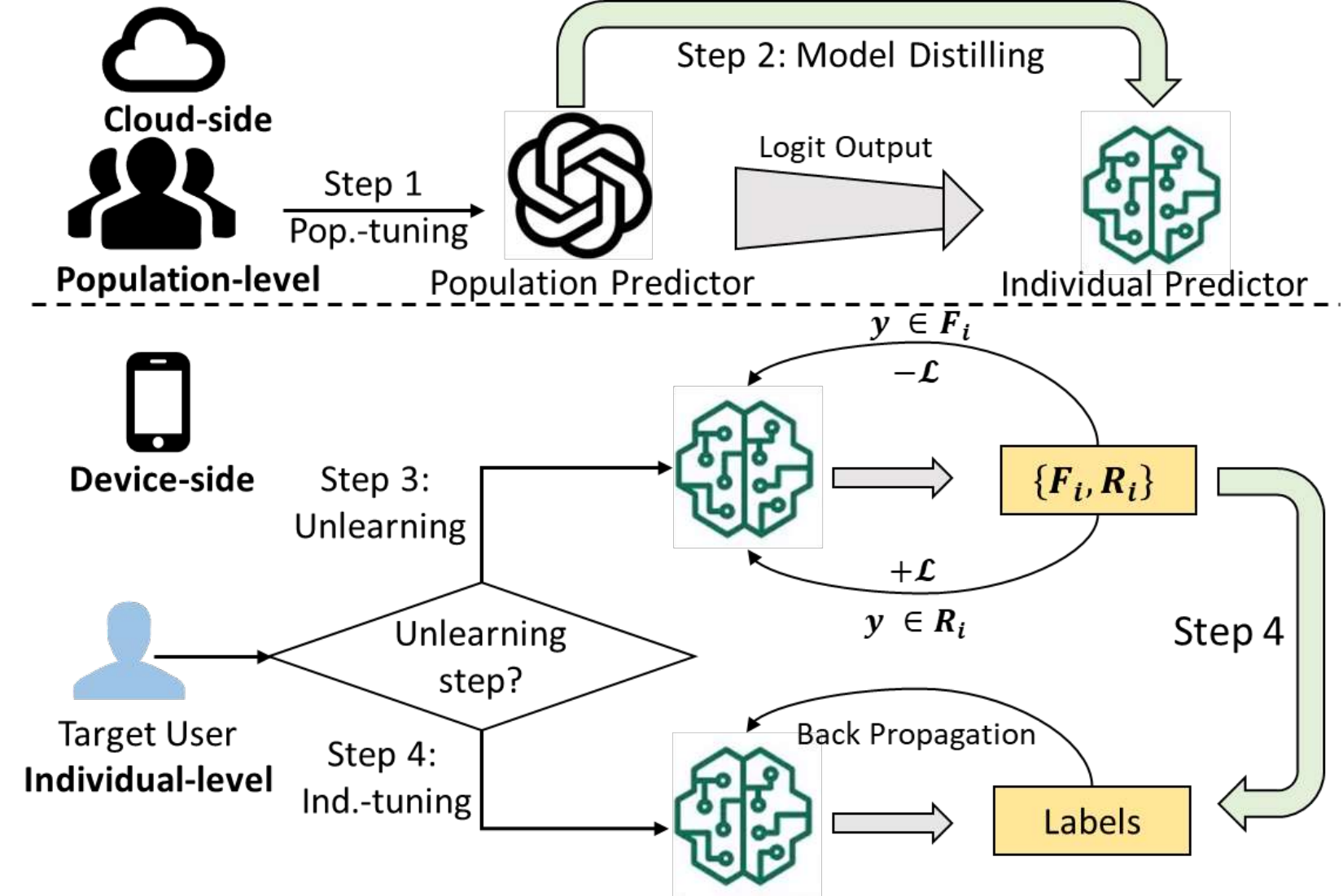}
    \vspace{-3mm}
    \caption{The workflow of PITuning framework.}
    \vspace{-1em}
    \label{fig:workflow}
    \end{figure} 
\section{method}

\begin{figure*}[t]
    \centering
    \includegraphics[width = 0.92\linewidth]{ 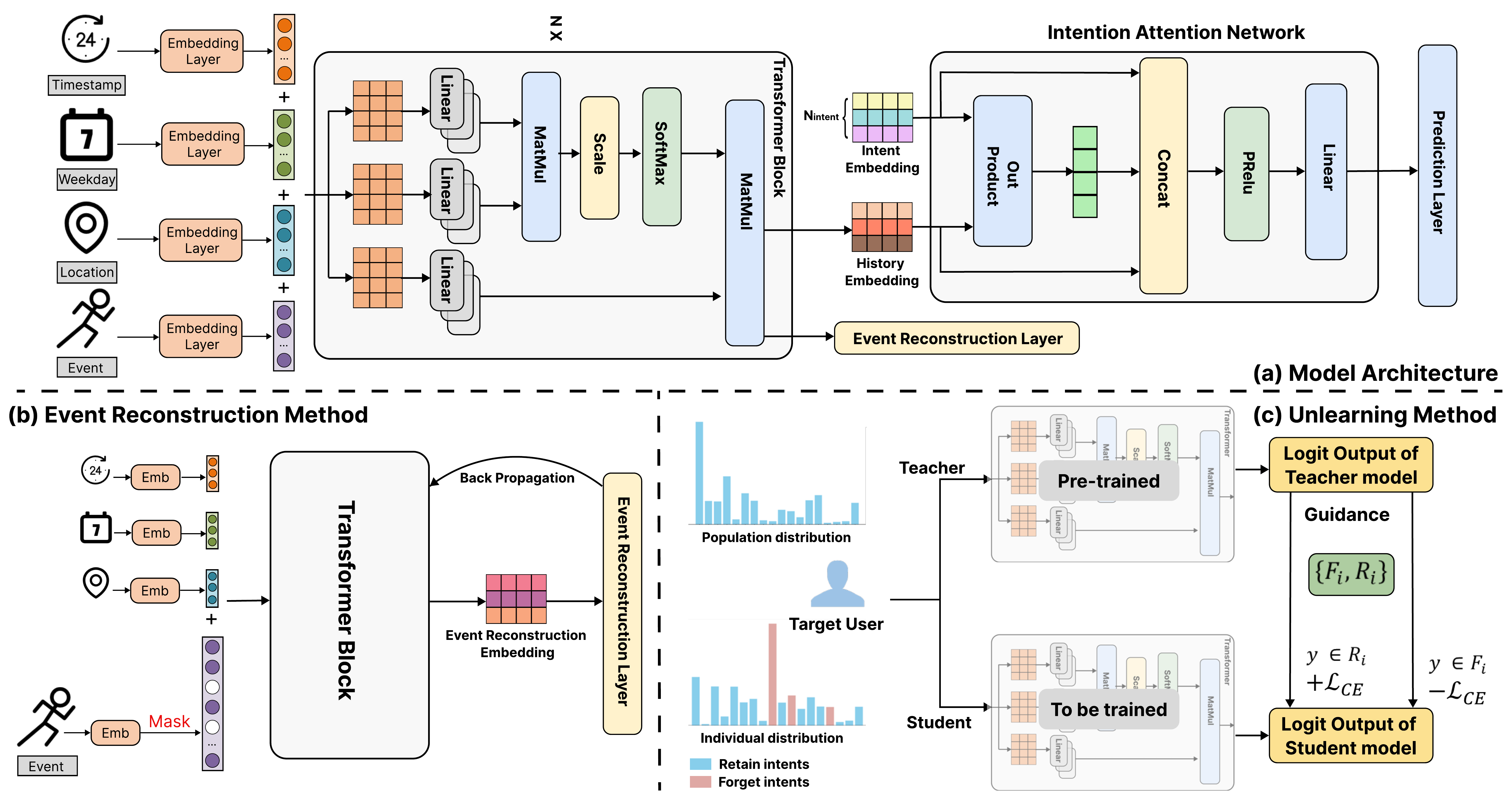}
    \vspace{-3mm}
    \caption{(a) The intent predictor architecture. (b) The masked event reconstruction in the population-level tuning. (c) The adaptive unlearning in the individual-level tuning.}
    \vspace{-1em}
    \label{fig:FM}
    \end{figure*} 

\subsection{Framework Overview}
We introduce our PITuning framework for adapting PLM to on-device intent prediction, as depicted in Figure \ref{fig:workflow}. On the cloud side, we utilize aggregated behavioral data collected from a population to fine-tune a global predictor, capturing population-level common behavioral patterns. Subsequently, we perform model distillation to obtain a lightweight predictor suitable for on-device deployment. On the device side, before further fine-tuning on individual data, we incorporate a novel unlearning strategy to identify and mitigate biases resulting from uneven learning of intents during population-level tuning. Finally, after two stages of PITuning, we attain a lightweight yet personalized model capable of accurate and efficient intent prediction on the device.
 
\subsection{Population-level Tuning}

We leverage population data alongside a PLM to model common behavioral patterns. The architecture of our model is depicted in Figure \ref{fig:FM}(a), where we integrate parameters from GPT2 \cite{GPT2}, an NLP pre-trained transformer model. Additionally, to enhance learning of event-to-intent transition patterns shared among the population, we introduce a masked event-reconstruction loss at the event level and utilize intent-wise attentive modeling atop the pretrained transformer. Finally, we distill a lightweight predictor under the guidance of the global predictor to meet deployment requirements on the device.

\subsubsection{Embedding layer.} Since we apply the NLP pre-trained model to a new modality, We create four embedding layers to get the location embedding $\textbf{E}_l \in \mathbb{R} ^{I \times d}$, weekday embedding $\textbf{E}_w \in \mathbb{R} ^{I \times d}$, time-slot embedding $\textbf{E}_t \in \mathbb{R} ^{I \times d}$ and event embedding $ \textbf{E}_e \in \mathbb{R} ^{I \times d} $ respectively, where $d$ denotes the embedding size.
\subsubsection{Transformer block.} In the global predictor, we employ the GPT2 model as the foundation for our transformer blocks. Trained on various web data, the GPT2 model is imbued with extensive knowledge, common sense, and fundamental principles, showing a robust capacity for generalization. We concat the location embedding, weekday embedding, time-slot embedding, and event embedding and put them into transformer blocks to obtain an implicit representation of the historical event sequence $H_t \in \mathbb{R} ^{I \times 4d}$, which can be formed as,
\begin{equation}\label{equ:TransBlock}
    \textbf{H}_t = \text{GPT2}(\text{concat}(\textbf{E}_l, \textbf{E}_w, \textbf{E}_t, \textbf{E}_e)). 
\end{equation}

\subsubsection{Intent-aware attentive modeling.} Notice that different intents exhibit preferences for varying lengths of historical data. To address this, we have developed a novel Intention Attention Network (IAT) that introduces a novel designed local activation unit to adaptively weigh sequences of historical events, accommodating the unique requirements of each intent. We create the learnable intent embedding matrix $\textbf{E}_i \in \mathbb{R} ^{N_I \times 4d}$ and feed it into IAT together with the history matrix.

Specifically, we apply activation units to the features derived from users' historical behaviors. These units function through a weighted sum pooling mechanism to adaptively compute the representation of intents, as detailed in Equation \ref{equ:IAT},
\begin{equation}\label{equ:IAT}
    \textbf{H}_w = \text{IAT}(\textbf{E}_i, \textbf{H}_t) = \sum_{j = 0}^{I} a(h_j,E_i)h_j = \sum_{j = 0}^{I} w_jh_j.
\end{equation}
Through this approach, $\textbf{H}_w$ changes across different intents, where $a(\cdot)$ represents a feed-forward network that yields activation weights. These weights are then combined through an outer product operation and integrated into the subsequent network layers to enhance relevance modeling.

Next, we use a Multilayer Perception (MLP) to be the prediction layer, which can be formed as,
\begin{equation}\label{equ:MLP}
\textbf{m} = f(\textbf{H}_w) = \mathbf{W}_2(\sigma (\mathbf{W}_1\textbf{H}_w+b_1))+b_2,
\end{equation}
where $\mathbf{W}, {b}$ are the trainable weight matrix and the bias matrix. The output of the MLP is the predicted intent distribution.

\subsubsection{Event-reconstruction auxiliary loss.}
To improve the model's proficiency in accurately capturing event-to-intent transition patterns, we employ a masked event reconstruction loss \cite{maskAutoEncoder}, which reconstructs the original event sequences based on the given partially observed signals, as shown in Figure \ref{fig:FM}(b).
Specifically, we randomly mask the event embedding $E_m$, and input them into the GPT2 model according to \ref{equ:TransBlock}. Next, we employ an MLP to be the event reconstruction layer to reconstruct the event sequence. The cross-entropy loss function is then used to assist model training. The loss function in population-level tuning can be formed as,
\begin{equation}\label{equ:poploss}
    \mathcal{L}_{pop} = \mathcal{L}_M + \mathcal{L}_{CE}(\textbf{m},L) = \mathcal{L}_{CE}(e_i, e_m) + \mathcal{L}_{CE}(\textbf{m},L), 
\end{equation}
where $e_i$ denotes the original event sequence, $e_m$ denotes the predicted event sequence, and $L$ denotes the ground truth of input.

\subsubsection{Model distilling.} To meet the requirement of deployment, we utilize the model distilling method, which is to train a smaller model (called the student model) to imitate the behavior of a larger model ( called the teacher model). The details of the model distillation process are shown in Appendix \ref{distilling}.
    
To guide the training of the student model, we design the soft loss for the soft targets, which is the Kullback-Leibler Divergence between the logit output of the teacher and the student network. Meanwhile, we also utilize the cross-entropy loss to ensure the student model learns the correct classifications. The loss function can be formed as follows,
\begin{align}\label{equ:distilling}
\mathcal{L}_{soft} = \text{KL}\left ( \frac{\textbf{M}_s}{\epsilon} ,\frac{\textbf{M}_t}{\epsilon} \right ), \;\;& \mathcal{L}_{hard} = \mathcal{L}_{CE}(\textbf{M}_s, L)  \\
 \mathcal{L}_D = \alpha  \mathcal{L}_{soft} &+\left ( 1-\alpha  \right ) \mathcal{L}_{hard}
\end{align}
where $\textbf{M}_t, \textbf{M}_s$ denotes the logits output of the teacher and student model respectively, and $\epsilon$ is a hyper-parameter, which means the temperature to smooth the probability distribution, while $\alpha$ is a hyper-parameter to balance the importance of two loss functions. By doing so, the student model learns both the fine-grained information from the teacher model's output and the essential classification ability, resulting in a smaller, more efficient model that retains much of the teacher model's predictive power. Subsequently, the student model is deployed on the device side.

\begin{algorithm} [h]
\caption{Population-to-Individual Tuning Framework} \label{algorithm}

\begin{algorithmic}[1]
\item[] \hspace*{-\algorithmicindent}\hspace*{-1.0ex} \large{\textbf{Population-level tuning}}
\Require Population data $x_i = \left ( l_i, w_i, t_i, e_i \right ) $
\Ensure The lightweight predictor $M_l$
\State $\textbf{E}_l, \textbf{E}_w, \textbf{E}_t, \textbf{E}_e \leftarrow emb(l_i), emb(w_i), emb(t_i), emb(e_i)$ \Comment{\textit{Input Embedding.}}
\State $\textbf{H}_t = \text{GPT2}(\text{concat}(\textbf{E}_l,\textbf{E}_w,\textbf{E}_t,\textbf{E}_e))$
\State $\textbf{M} = \text{MLP}(\text{IAT}(\textbf{H}_t))$ \Comment{\textit{Population Intent Prediction.}}
\State $\textbf{E}_m = \text{Mask}(\textbf{E}_e)$ \Comment{\textit{Mask the event sequence. }}
\State $\textbf{H}_m = \text{GPT2}( \text{concat}(\textbf{E}_l, \textbf{E}_w, \textbf{E}_t, \textbf{E}_m))$
\State $\textbf{M}_m = \text{MLP}(\textbf{H}_t)$ \Comment{\textit{Event Reconstruction.}}
\State Model Distillation to obtain the lightweight predictor $M_l$. 
\end{algorithmic}

\setcounter{ALG@line}{0} 

\begin{algorithmic}[1]
\item[] \hspace*{-\algorithmicindent}\hspace*{-1.0ex} \large{\textbf{Individual-level tuning}}
\Require Individual data $x_j= \left ( l_j, w_j, t_j, e_j \right )$ of User $j$
\Ensure The tuned lightweight predictor $M_f$
\State $(F_i, R_i) \gets \text{ManageIntents}(P_{pop}, P_{in})$
\If{$F_j \ne \varnothing$}
\State the adaptive unlearning to forget $F_j$
\EndIf
\State Finetune the lightweight predictor $M_f$
\end{algorithmic}

\end{algorithm}

\subsection{Individual-level Tuning}
In the individual-level tuning, we harness personalized user data to fine-tune the model, enabling it to adapt to and learn individual preferences. However, the key challenge is that there may be a significant difference between population intent distribution and individual intent distribution~(as shown in our previous data analysis in Figure~\ref{fig:tsne}), leading to a bias, particularly for some long-tail intents during the population tuning stage. Therefore we first design an adaptive unlearning strategy to help the model disregard these biases. After that, we can finetune a personalized model that is both accurate and efficient.

\subsubsection{Adaptive unlearning on biased intents.} Unlearning involves intentionally disregarding or ignoring specific data or patterns in a trained neural network \cite{Unlearning}. Initially, we decide whether each user's intent should be forgotten or retained. We propose two methods to identify the forgotten intents. First, we analyze the intent distribution of the global predictor output at the population level. If the proportion of an intent $P_{out}(i)$ is less than the threshold $\varepsilon$, it is part of the static forgotten set $F_{sta}$,

\begin{equation}\label{equ:forgotton_sta}
    I_i \in F_{sta} \;\; \text{if} \;\; P_{out}(I_i) < \varepsilon,
\end{equation}.
where $I_i$ denotes the intents $i$. Secondly, if the proportion of intent at the population level $P_(pop)$ is less than the average while the proportion at the individual level $P_(in)$ is greater than the average, then it belongs to the dynamic forgotten set $F_{dyn}$,
\begin{equation}\label{equ:forgotton_fyn}
    I_i \in F_{dyn} \;\; \text{if} \;\;P_{pop} < \frac{1}{N_I} \;\;\text{and}\;\; P_{in} > \frac{1}{N_I},
\end{equation}
Finally, the union of static forgotten set and dynamic forgotten set is taken as the forgotten set, which can be formed as,
\begin{equation}\label{equ:intentselection}
    F_i \;=\; F_{dyn} \cup F_{sta},\; R_i \;= \;\text{others},
\end{equation}.

 where $F_i, R_i$ represents the forgotten set and retained set of user $i$. If it is determined that intent needs to be forgotten by the user, the adaptive unlearning is applied; if not, it is deemed unnecessary.
To effectively achieve the unlearning goal, we design an unlearning loss:
\begin{equation}\label{equ:unlearningLoss}
\mathcal{L}_{un} = \lambda \mathcal{L}_{CE}( \textbf{M}_r, R_i) - \mathcal{L}_{CE}( \textbf{M}_f, F_i),
\end{equation}
where $\textbf{M}_f$ signifies the model's output for the intent category designated for forgetting, and $\lambda$ is a hyper-parameter to balance the trade-off between forgetting and retaining. Intuitively, during the unlearning process, the model is learned to minimize the loss between the output from the updated model and the original model on the intent to retain while maximizing the loss between the output from them on the data to forget.

\subsubsection{Finetuning for personalized model.} Finally, we utilize personalized individual data to fine-tune the model for each user. This fine-tuning process enables the model to transition from capturing common behavioral patterns to reflecting a user's unique preferences, thereby enhancing the accuracy of the model's predictions. We also use cross-entropy loss to guide model tuning.


\section{experiment}
\subsection{Experiment Settings}
\subsubsection{Datasets.}

We evaluate the performance of our model on two large-scale real-world activity datasets.
\begin{packed_itemize}
\item \textbf{Honor Dataset.} 
The Honor Dataset is sampled from the usage log of the mobile phones. When a user uses mobile phones, various types of logs are generated, desensitized and reported (with user consent). We selected 114 types of events that are commonly monitored in most mobile applications and classified them into 18 intents, which cover the aspects of news, study, work, entertainment, sports, etc. We sampled two datasets between June 1st and August 22nd, 2023 (the first) and August 22nd and September 10th, 2023 (the second) which in total contain 4,500 and 5,000 anonymous users.

\item \textbf{Mobile Dataset.}
The Mobile Dataset consists of anonymous user trajectory data collected by a major mobile network operator in China in October. The dataset comprises 6,000 users, of which, at the population level, we select 4,000 users for training, and at the individual level, we select the remaining users. In this dataset, we use the location category as the activity and intent type.  
\end{packed_itemize}
Table \ref{tab:datasets} shows the statistics of the Honor dataset and Mobile dataset. The large-scale and fine-grained datasets can ensure the validity of the model test.
\begin{table}[t]
\centering
\small
\caption{Statistics of the datasets used in our experiments. }
{
\begin{tabular}{c | c| c c } 
 \toprule
 \multicolumn{2}{c}{Datasets} & \textbf{Honor Dataset} & \textbf{Mobile Dataset} \\
 \midrule

 \multicolumn{2}{c}{Type of Events} & 114 & 12 \\
 \multicolumn{2}{c}{Type of Intents} & 18 & 12 \\

 \midrule
\multirow{3}*{\makecell[c]{Population \\ level}} & Users &4,500 & 4,000 \\
&Duration & 6.1-8.22, 2023 & 10.1-31, 2016 \\
&Number of logs & 10,376,148 & 334,651 \\
\midrule
\multirow{3}*{\makecell[c]{Individual \\ level}} & Users &5,000 & 2,000 \\
& Duration & 8.23-9.10, 2023 & 10.1-31, 2016 \\
&Number of logs & 976,788 & 208,161 \\
  \bottomrule
 
\end{tabular}}
\vspace{-2em}
\label{tab:datasets}
\end{table}

\begin{table*}[t]
\centering
\caption{Overall prediction performance PITuning compared with baselines on Honor and Mobile datasets. }
\scalebox{1.0}{
\begin{tabular}{c| c c c c c c | c c c c c c} 
 \toprule
  & \multicolumn{6}{c}{\textbf{Honor Dataset}} & \multicolumn{6}{|c}{\textbf{Mobile Dataset}} \\
 \midrule
 \textbf{Model} &$Prec_w$ &$Rec_w$ &$Prec_m$ &$Rec_m$&N@3&N@5&$Prec_w$ &$Rec_w$ &$Prec_m$ &$Rec_m$&N@3&N@5\\
 \midrule
 CLOVER &0.4479 &0.4516 &0.2494 &0.2527 &0.6094&0.6213 &0.7052 &0.7505 &0.6004 &0.6350 &0.7860 &0.8270 \\
 MetaBert4Rec &0.4683 &0.5028 &0.2888 &0.3218   &0.6842 &0.7023 &0.7714 &0.8174 &0.6419 &0.6739 &0.8268 &0.8814\\
 \midrule
 P5  &0.4722 &0.5161 &0.2452 &0.2807  &0.6045 &0.6284  &0.7436 &0.7930 &0.6133 &0.6410 &0.8104 &0.8672\\
 InstructRec  &0.4315 &0.4680 &0.2423 &0.2684   &0.6407 &0.6724  &0.7318 &0.7743 &0.6043 &0.6397 &0.7968 &0.8444\\
 LSAT  &0.4714 &0.5008 &0.2778 &0.2983 &0.6055 &0.6301   &0.7572 &0.8058 &0.6415 &0.6845 &0.8220 &0.8809\\
 OFA  &0.4928 &\underline{0.5243} &0.3366 &0.3756  &0.7032 &0.7244  &\underline{0.7851} &0.8258 &0.6538 &0.7062 &0.8488 &0.9024\\
 TallRec  &0.4486 &0.4764 &0.2659 &0.3083  &0.5972 &0.6102  &0.7200 &0.7693 &0.6261 &0.6532 &0.7949 &0.8345\\
 \midrule
 EODRec  &0.4517 &0.4958 &0.2799 &0.2903  &0.5867 &0.6039  &0.7489 &0.8087 &0.6072 &0.6519 &0.8296 &0.8939\\
 MPDA  &\underline{0.4947} &0.5197 &\underline{0.3408} &\underline{0.3841}  &\underline{0.7117} &\underline{0.7371}  &0.7785 &\underline{0.8361} &\underline{0.6581} &\underline{0.7085} &\underline{0.8542} &\underline{0.9185}\\
 \midrule
 \textbf{ours}  &\textbf{0.5374} &\textbf{0.5599} &\textbf{0.4693 }&\textbf{0.4840 } &\textbf{0.7329} &\textbf{0.7626}   &\textbf{0.8715 }&\textbf{0.9002} &\textbf{0.8449} &\textbf{0.8802} &\textbf{0.9506} &\textbf{0.9537}\\
 Improv.  &8.63\%	 &6.79\%	 &37.71\%	 &26.01\%	 &2.98\%	 &3.46\%	 &11.00\%	 &7.67\%	 &28.38\%	 &24.23\%	 &11.29\%	 &3.83\%
 \\

 \bottomrule
\end{tabular}}
\label{tab:overall}
\end{table*}

\subsubsection{Metrics.} To assess model performance, we employ five widely used metrics: weighted precision ($Prec_w$), weighted recall ($Rec_w$), macro precision ($Prec_m$), macro recall ($Rec_m$), and NDCG(N). Weighted metrics and NDCG gauge classification accuracy and ranking quality, respectively, while macro metrics evaluate the average prediction accuracy for each intent, indicating the model's predictive quality across intents. A smaller gap between weighted and macro metrics implies consistent prediction accuracy across intents, reflecting fairness. Conversely, a large gap suggests inadequate modeling of long-tail intents, leading to suboptimal outcomes. Refer to Appendix \ref{metrics} for metric calculations.

\subsubsection{Baselines.} 
We elaborately select the following nine representatives to be compared with our proposed algorithms, which cover the meta-learning methods for personalized recommendations (CLOVER~\cite{Wei22KDDCLOVER}, MetaBert4Rec~\cite{Kim23MetaBert4Rec}), LLM-based recommendations (P5~\cite{Geng22P5}, InstructRec~\cite{Zhang2023Insprom}, LSAT~\cite{Shi2023LSAT}, One fits All (OFA)~\cite{zhou2023onefitsall}, TallRec~\cite{Bao23Tallrec}) and device-cloud collaboration recommendations (EODRec~\cite{Xia23DeviceSession}, MPDA~\cite{yan2022devicebase}). We provide the details of baselines in Appendix \ref{baseline}.

\subsubsection{Implementation Details.}
Our model employs the Adam optimizer with a learning rate of 0.01 across two tuning phases. And we set the input length as 30. During the population-level tuning stage, we utilize the GPT2 small version for the transformer blocks, which features a 12-layer transformer architecture and a 768-dimensional feature space, while on the individual-level tuning stage, we choose a transformer decoder with 4-layer and a 768-dimensional feature space to meet the deployment requirement. As for hyper-parameters $\alpha$, $\epsilon$, and $\lambda$, we set 0.5, 1, and 1 respectively. Details of hyperparameters are shown in Appendix \ref{hp}. 
In the honor dataset, one week is allocated for training, one day for validation, and four days for testing. In the mobile dataset, 60\% of the data is used for training, with 10\% for validation and 30\% for testing. The code is available at \href{https://github.com/tsinghua-fib-lab/LLM-for-User-Intent}{https://github.com/tsinghua-fib-lab/LLM-for-User-Intent}.

\subsection{Overall Performance}

In Table \ref{tab:overall}, we display the overall results of our model, meta-learning methods (CLOVER, MetaBert4Rec), LLM-based recommendations (P5, InstructRec, LSAT, OFA, TallRec) and device-cloud collaboration recommendations (EODRec, MPDA) to predict the next user intention in two datasets. We list three metrics of all methods. From the result, we have the following findings:

\begin{packed_itemize}

\item \textbf{Our framework steadily achieves the best performance.} Our model gets superior results on both datasets and performs better than other compared algorithms. For example, the macro metrics improvement of our model is around 24\% to 37\% compared with the second-best performance model (MPDA). The $NDCG$ improvement of our model is about 3\% to 11\%.

\item \textbf{Our model has the smallest difference between weighted metrics and macro metrics.} Weighted metrics and macro metrics count the global accuracy and the average accuracy of each intent respectively. A smaller difference suggests comparable prediction accuracy across different intents, indicating fairness among the intents. Our model utilizes adaptive unlearning to effectively correct the long-tail intent learning bias caused by the model in population-level tuning, and improve the accuracy.

\item \textbf{The method of LLM-based model with device-cloud collaboration is necessary for user behavior modeling.} MPDA utilizes the LLM-based model with the device-cloud collaboration method resulting in the best performance within the baseline. However, MPDA fails to fully exploit the benefits of diverse user data available on the cloud side. In contrast, our proposed PITuning framework captures generalized user behavior patterns, leading to superior performance.


\end{packed_itemize}

\subsection{Ablation Study}



\begin{figure}[t]
\centering
\subcaptionbox{Honor dataset.\label{fig:ablation_honor}}{
    \includegraphics[width=0.46\linewidth]{ 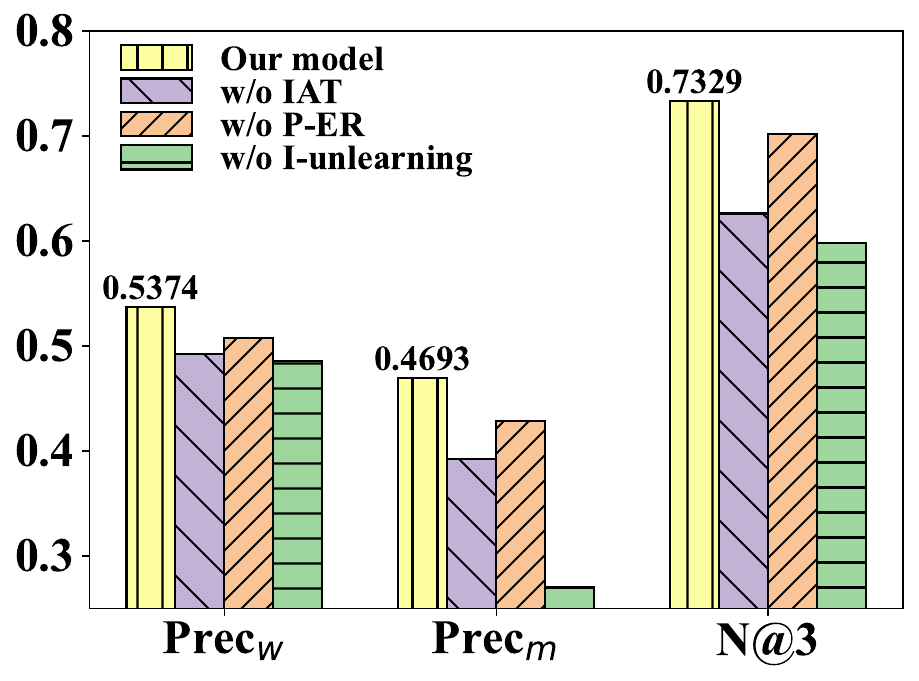 }}
\subcaptionbox{Moblie dataset.\label{fig:ablation_mobile}}{
    \includegraphics[width=0.46\linewidth]{ 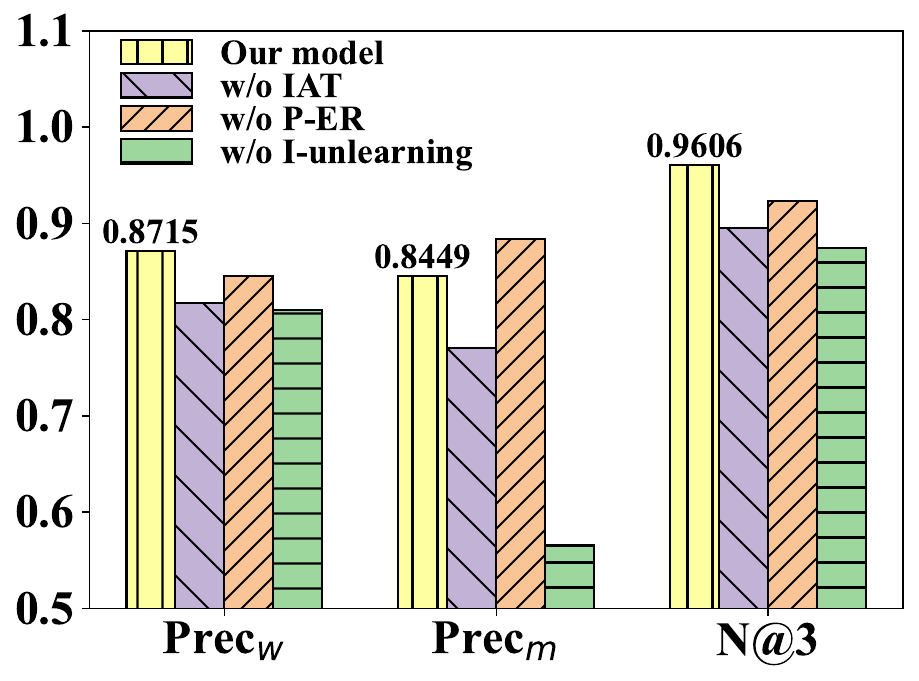}}
    \vspace{-3mm}
\caption{Ablation study.}
\label{fig:ablation}
\vspace{-3mm}
\end{figure}

To gain a deeper understanding of each component of our model, we carried out a sequence of ablation studies. Firstly, we removed the Intention Attention Network (IAT) within this model, followed by removing the event reconstruction loss in the population-level tuning (P-ER). Subsequently, we removed the adaptive unlearning in the individual-level tuning(I-unlearning).

The results of the ablation study are presented in Figure \ref{fig:ablation}.  We observed that the absence of the Intention Attention Network (IAT) hindered the model's ability to appropriately assign weights to each intent, consequently impacting both $Prec_w$ and $Prec_m$. Additionally, the event reconstruction loss played a pivotal role in guiding the transformer block towards more accurate modelling of users' historical event sequences, thereby enhancing the model's performance.
Furthermore, we noticed that the omission of the adaptive unlearning compromised the model's capacity to effectively handle long-tail intents, resulting in a significant reduction in $Prec_m$ by approximately 37\%. 

\subsection{Analysis of Population-level Tuning} 

$\bullet$ \textbf{Performance improvement brought by pretrained LM.}

\begin{figure}[t]
\centering
\subcaptionbox{Prediction accuracy\label{fig:case_study_llm_acc}}{
    \includegraphics[width=0.46\linewidth]{ 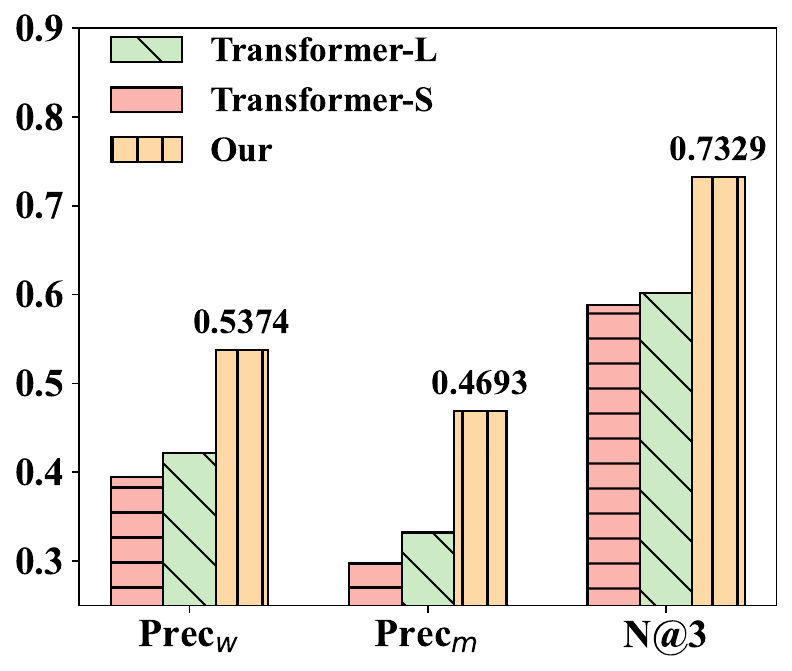 }}
\subcaptionbox{Training convergence\label{fig:case_study_LLM_loss}}{
    \includegraphics[width=0.46\linewidth]{ 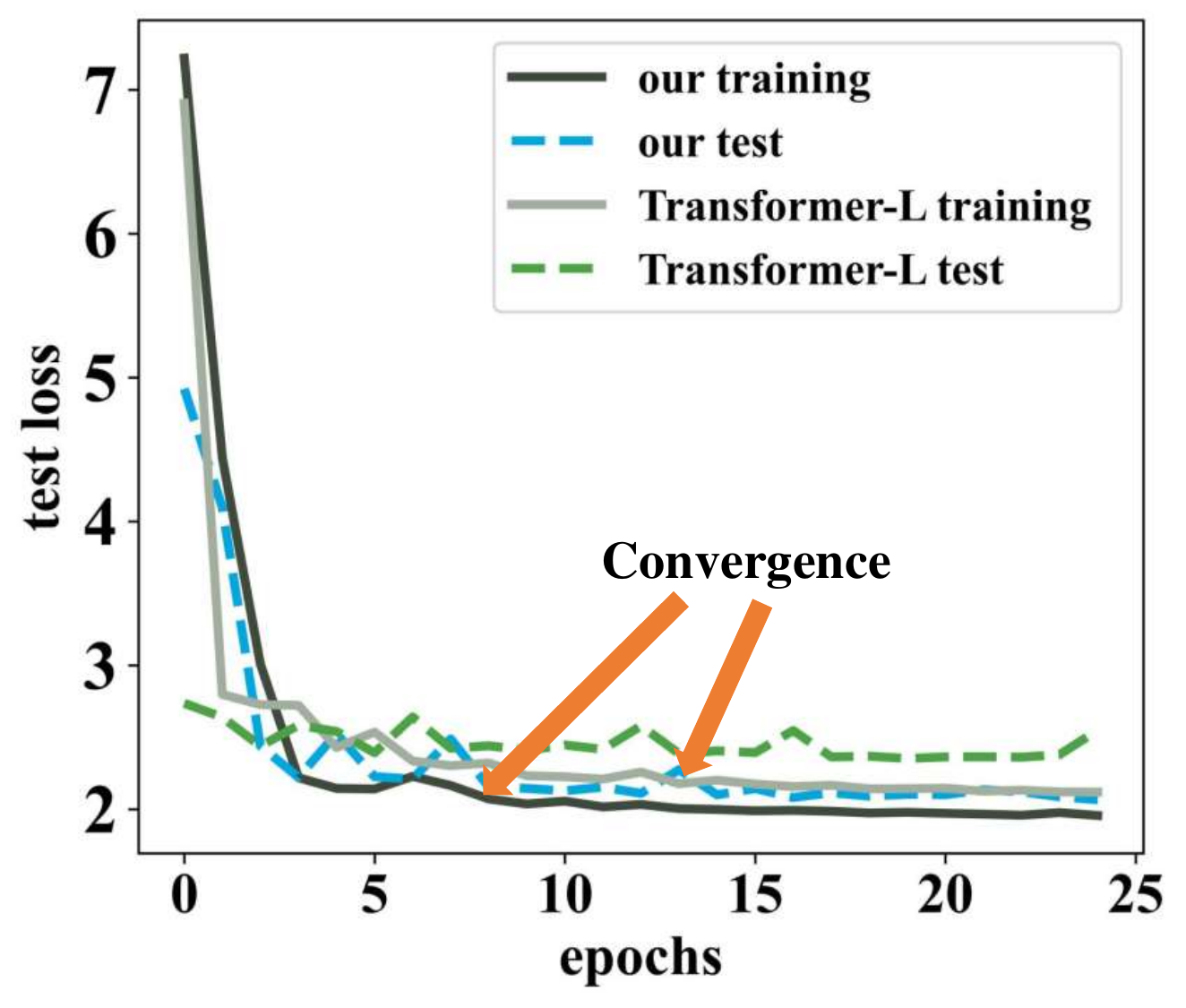}}
    \vspace{-3mm}
\caption{Comparing performance without pretrained LM.}
\label{fig:case_study_llm}
\vspace{-3mm}
\end{figure}

We assessed how LLM contributes to modeling population-level common behavior patterns by replacing the LLM with transformer encoders of two sizes. One matches the size of GPT2 (Transformer-L), while the other matches the size of the lightweight predictor we distilled (Transformer-S), allowing them to train from scratch.

Figure \ref{fig:case_study_llm} compares the prediction accuracy and loss among the models. Our analysis underscores that without leveraging the PTM, the model lacked foundational common sense and rule-based guidance, leading to a significant decline in its ability to capture common behavioral patterns and accuracy. Additionally, the Transformer-S, with fewer parameters, encountered challenges in modeling complex user behaviors, resulting in inferior performance. Moreover, with the guidance of LLM, the model demonstrated faster convergence.

$\bullet$ \textbf{Effectiveness of extracting common behavioral pattern.} 

To highlight the effectiveness of IAT in capturing intent-aware transition patterns in user event sequences, we visualize the attention map between intents and historical sequences. Figure \ref{fig:att_map} presents the resulting attention maps, highlighting the IAT's ability to discern each intent's preference for historical sequence length. Analysis of the attention map reveals that certain intents, such as short video, game, and photo intents, predominantly rely on short-term historical sequences. Conversely, intents like checking the weather, taking a taxi, and exercising necessitate long-term historical sequences. Additionally, some intents rely on both short-term and long-term historical sequences, such as checking the weather, music, and audiobook intents. These insights uncover users' daily behavior patterns, enabling researchers to construct more nuanced historical sequences and features to enhance accuracy.
\begin{figure}[tb]
\vspace{-3mm}
    \centering
    \includegraphics[width = 0.92\linewidth]{ 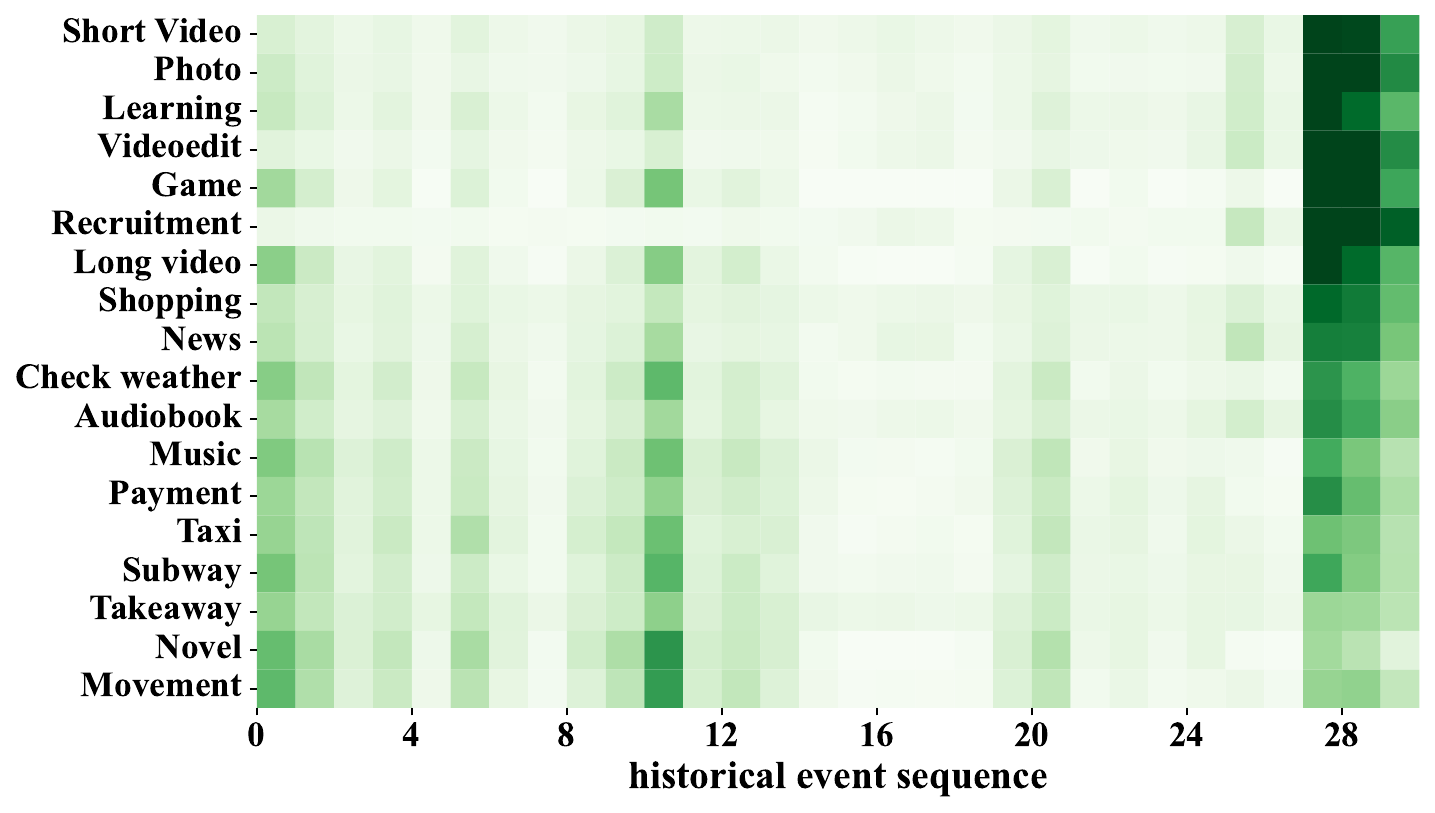}
    \vspace{-3mm}
    \caption{The attention map of different intents illustrating diverse event-to-intent transition patterns.}
    \label{fig:att_map}
    \end{figure} 

\begin{figure}[t]
\centering
\subcaptionbox{PITuning vs. groundtruth.\label{fig:event_our}}{
    \includegraphics[width=0.46\linewidth]{ 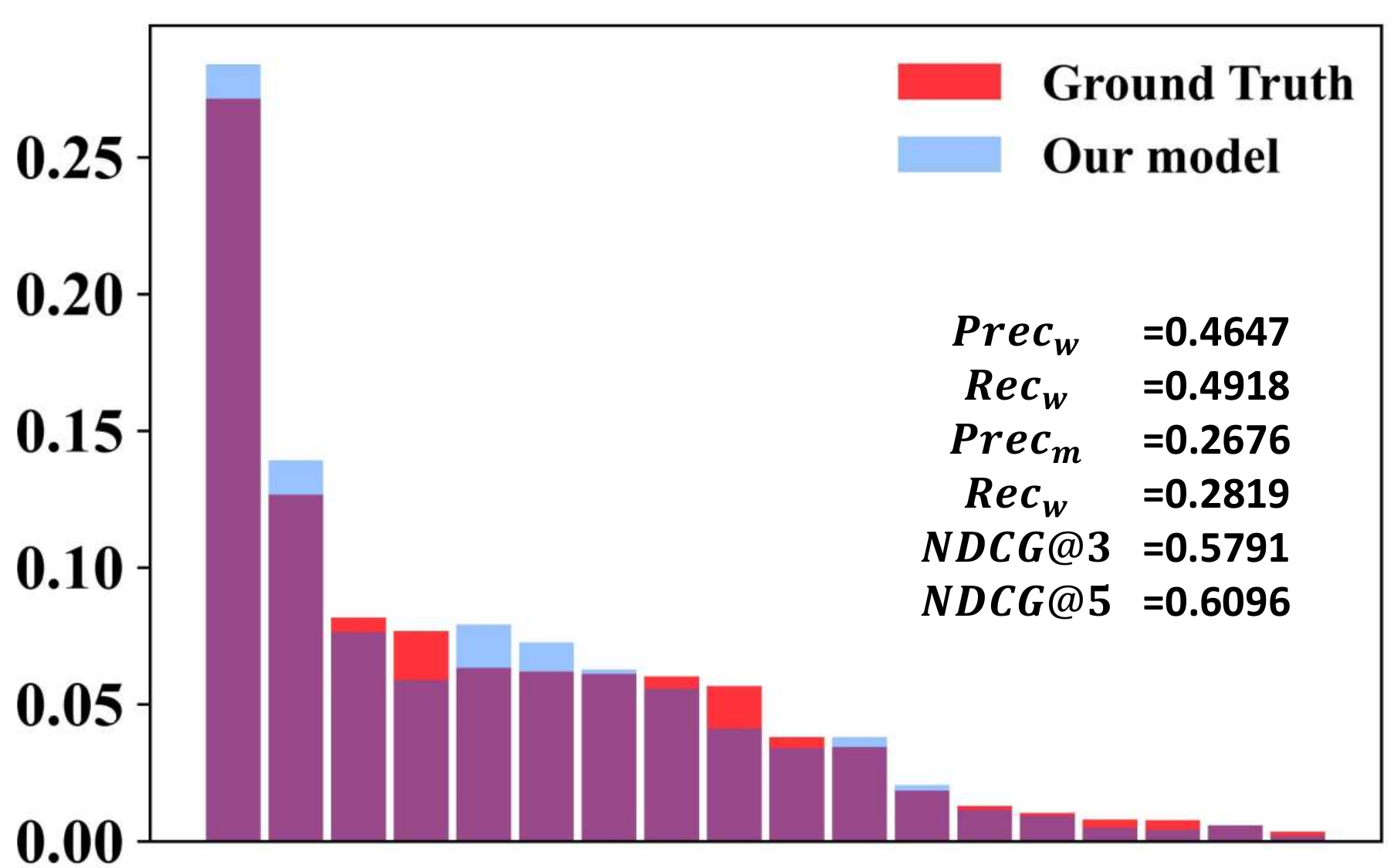 }}
\subcaptionbox{OFA vs. groundtruth.\label{fig:event_ofa}}{
    \includegraphics[width=0.46\linewidth]{ 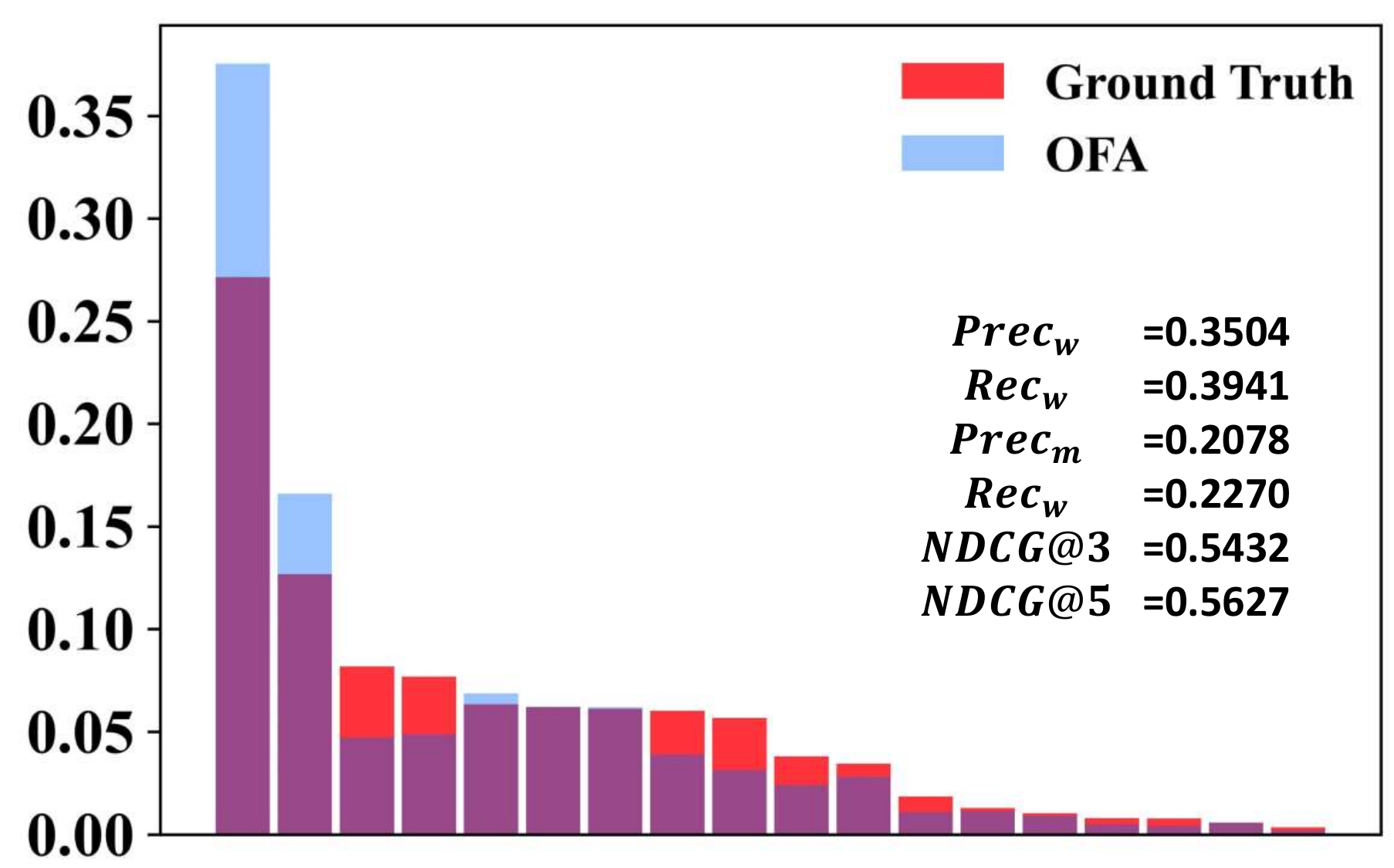}}
    \vspace{-3mm}
\caption{Comparison of intent distribution generated by PITuning and OFA after population-level tuning.}
\label{fig:event}
\vspace{-3mm}
\end{figure}

To showcase the effectiveness of the event reconstruction loss, we compare the difference between the intent distribution output by our model and OFA during the population-level tuning stage. Figure \ref{fig:event} presents the results, demonstrating that our model could well model transitions from events to intents with the help of the event reconstruction loss. The closer the output intent distribution is to the real distribution, the more conducive it is to capture the common behavior patterns in the population-level tuning.

\subsection{Analysis of Individual-level Tuning.}
\ 
\newline
$\bullet$ \textbf{Choice of device model.} 

To evaluate the efficiency of the lightweight predictor obtained through model distillation, we compared it with the tree model LightGBM \cite{ke2017lightgbm}, and two variants of the original population predictor at the cloud side, i.e., full-parameter tuned population predictor (FP), and partial parameter tuned population predictor (PP). Additional details about the tree model can be found in Appendix \ref{treemodel}. PP is inspired by the cross-domain adaptation techniques used in OFA \cite{zhou2023onefitsall}, which argues that self-attention layers and feed-forward neural networks encapsulate most learned knowledge and can be frozen during the finetuning process.

The device model results, shown in Table \ref{tab:distilling}, indicate that the full-parameter tuning method achieves higher performance. However, its large parameter size makes it challenging to implement on the device side. The partial parameter method struggles to transfer individual preferences from population-level common preferences, resulting in lower performance. Although LGBM has a smaller parameter count, its stability is inferior, leading to decreased accuracy.

\begin{table}[t]
\centering
\caption{Prediction performance using different model structures on the device side. }
\scalebox{0.82}{
\begin{tabular}{c| c c  | c  c | c c} 
 \toprule
  \multirow{2}*{\textbf{Model}} & \multicolumn{2}{c|}{\textbf{Honor}} & \multicolumn{2}{c|}{\textbf{Mobile}}  & \multirow{2}*{\textbf{\makecell[c]{Infer \\ speed}}}& \multirow{2}*{\textbf{\makecell[c]{Params\\(Trainable) \\ }}} \\
  & $Prec_w$ & $Prec_m$ &$Prec_w$ & $Prec_m$& \\
 \midrule
 FP &\textbf{0.5496 }&\textbf{0.4768} &\textbf{0.8803} &\textbf{0.8524} &8.9ms & 138M(138M) \\
PP &0.5217 &0.4528 &0.8659 &0.8322 &8.9ms &138M (39M) \\
 LGBM  &0.4742 &0.3586 &0.7943 &0.6485 &\textbf{15us} &\textbf{8.8K(8.8K)}  \\
 \midrule
 \textbf{Our}  &\underline{0.5374 }&\underline{0.4693} &\underline{0.8715} &\underline{0.8449} &\underline{2.85ms} &\underline{10.86M(10.86M)}  \\
 \bottomrule
\end{tabular}}
\label{tab:distilling}
\end{table}

$\bullet$ \textbf{Effectiveness of adaptive unlearning.} 



\begin{figure}[t]
\centering
\subcaptionbox{Honor Dataset.\label{fig:unlearning_honor}}{
    \includegraphics[width=0.46\linewidth]{ 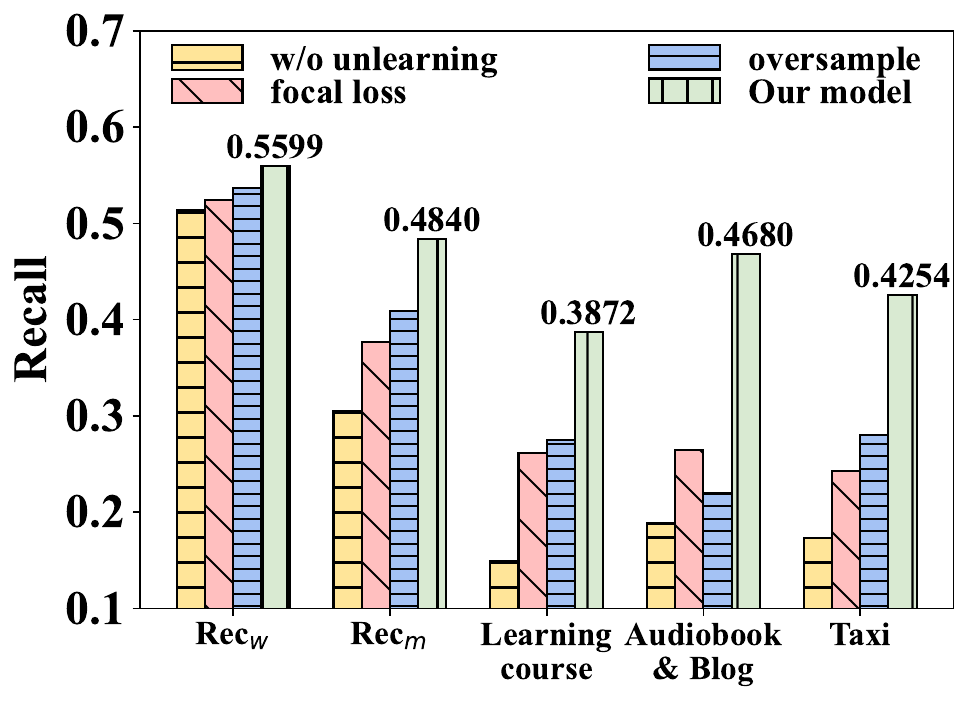 }}
\subcaptionbox{Mobile Dataset.\label{fig:unlearning_mobile}}{
    \includegraphics[width=0.46\linewidth]{ 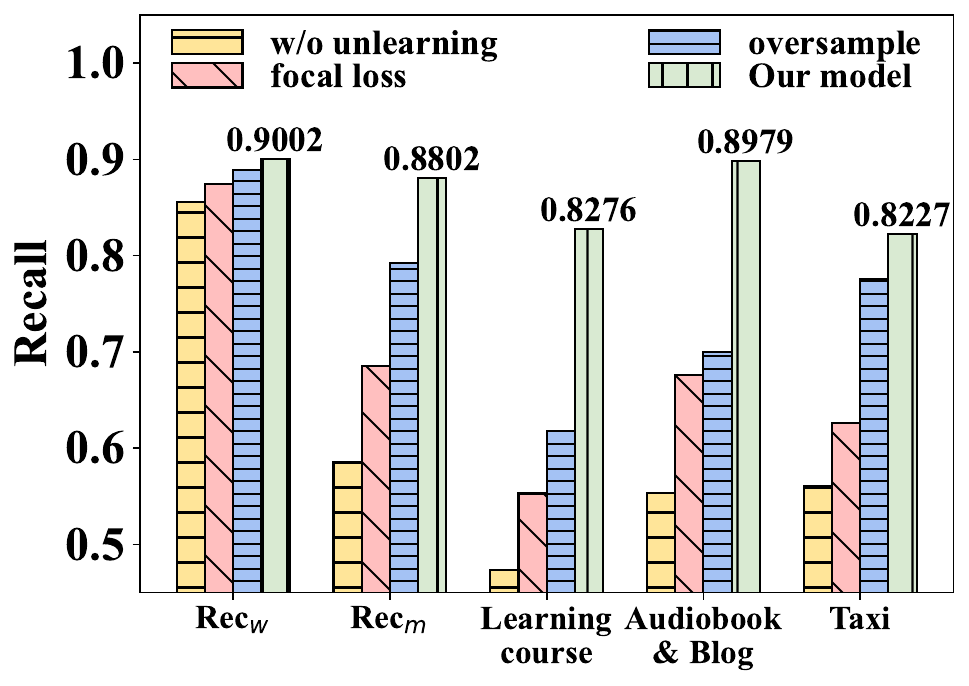}}
    \vspace{-3mm}
\caption{effectiveness of adaptive unlearning strategy compared with other class imbalance handling methods.}
\label{fig:unlearning}
\vspace{-3mm}
\end{figure}
To demonstrate the effectiveness of adaptive unlearning in enhancing the accuracy of long-tail intents, we compared it with oversampling methods and focal loss \cite{Lin2017focalloss}. We focused on the three intents with the smallest proportions in the datasets and evaluated recall, which effectively reflects the model's performance in identifying long-tail intents.

The results, shown in Figure \ref{fig:unlearning}, indicate although focal loss and oversampling methods show some improvement, their $Rec_w$ and $Rec_m$ still differ, indicating they fail to address the deviation caused by the disparity in intention distribution between population and individual levels. Through adaptive unlearning, the model gradually overcomes biases towards these long-tail intents in population-level tuning, resulting in significant improvements in precision and recall.


\subsection{Practicability Study}

$\bullet$ \textbf{Sensitivity of individual data scale in individual-level tuning.} 

In the individual-level tuning stage, particularly on the device side, there are limitations in storage and computing resources. To investigate the impact of data size, we conducted experiments by varying the data size in individual-level tuning and compared it with the second-best model (MPDA).

The results, shown in Figure \ref{fig:datasize}, indicate that increasing the dataset size leads to marginal performance enhancements across all models. This trend highlights our model's capability to capture user behavior preferences. However, larger datasets significantly increase demands for computing power and storage resources. Therefore, to strike a balance between model effectiveness and computational efficiency, we selected a one-week dataset size.
\begin{figure}[t]
\centering
\subcaptionbox{Precision of Honor Dataset.\label{fig:datasize_prec}}{
    \includegraphics[width=0.46\linewidth]{ 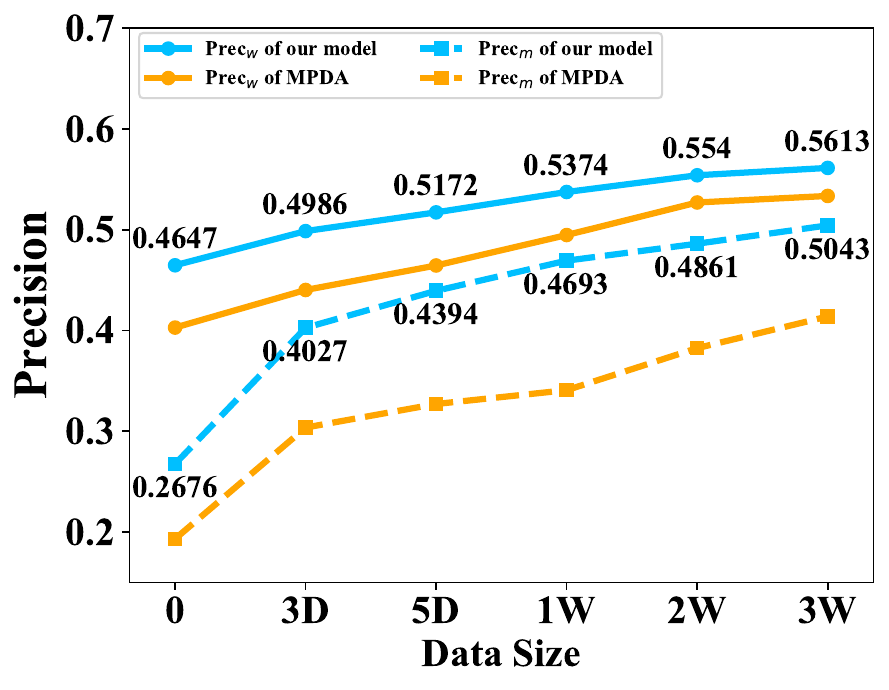 }}
\subcaptionbox{Recall of Honor Dataset.\label{fig:datasize_rec}}{
    \includegraphics[width=0.46\linewidth]{ 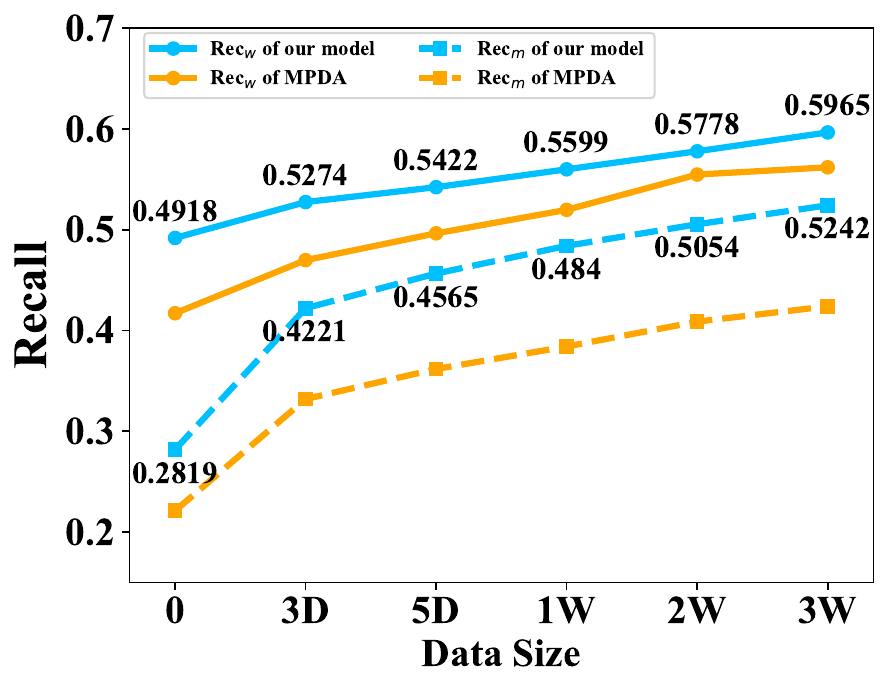}}
    \vspace{-3mm}
\caption{Influence of individual data size on performance.}
\label{fig:datasize}
\vspace{-3mm}
\end{figure}

$\bullet$ \textbf{sensitivity of event sequence length.}

To investigate the influence of the event sequence length, we conduct experiments by changing the input length of the historical series, and compared with the second-best model (MPDA). The results, illustrated in Figure \ref{fig:length}, show a slight improvement in performance across all models with increasing input length. This trend highlights our model's ability to capture long-term dependencies. However, longer input lengths substantially increase computational demands. Thus, to balance model performance and computational efficiency, we selected an input length of 30.

\begin{figure}[t]
\centering
\subcaptionbox{Precision of Honor Dataset.\label{fig:length_honor}}{
    \includegraphics[width=0.46\linewidth]{ 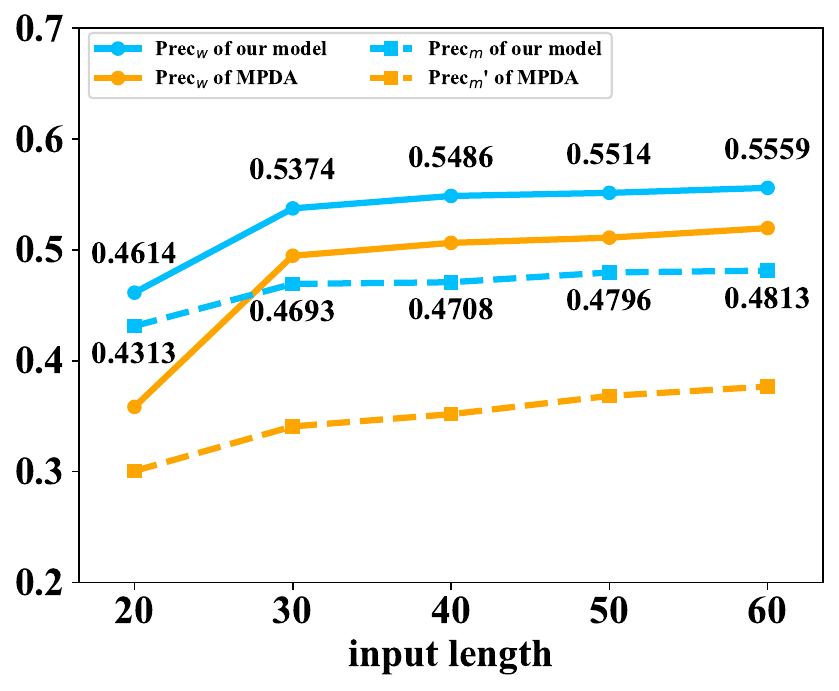 }}
\subcaptionbox{Precision of Mobile Dataset.\label{fig:length_mobile}}{
    \includegraphics[width=0.46\linewidth]{ 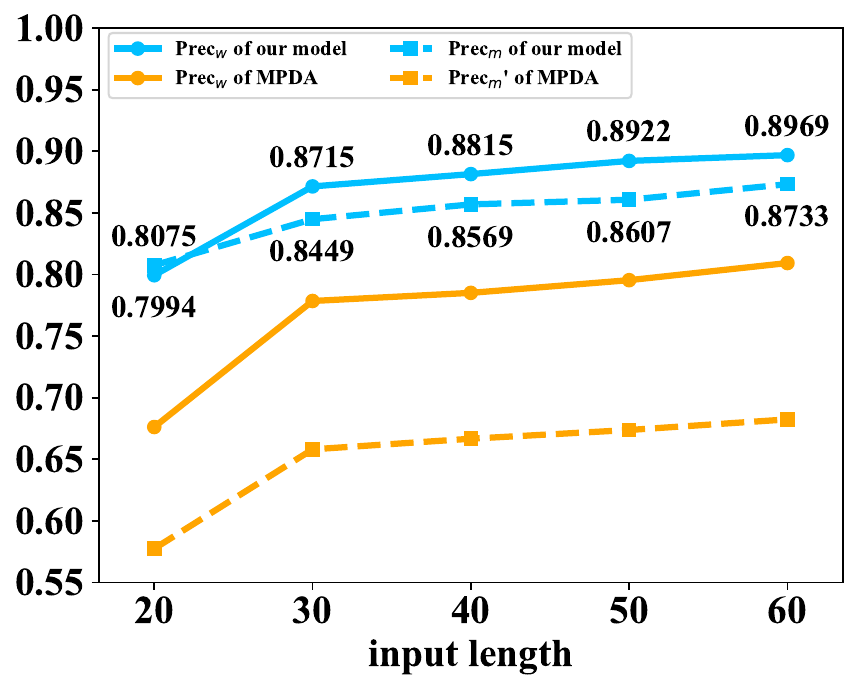}}
    \vspace{-3mm}
\caption{Influence of event seq. length on performance.}
\label{fig:length}
\vspace{-3mm}
\end{figure}


\section{related works}
\subsection{User Intent Prediction}

User intent prediction model, a recommendation system, emphasize modeling user-event interaction sequences. Recent works \cite{Pu19transformer, Wang2021transformer} integrate transformers into various models. \citet{yang20iart} introduce intent-aware ranking with transformers, incorporating intent-aware utterance attention. Meanwhile, \citet{wang21masked} propose a masked-field framework for distinct representations per intent.

Recent advancements \cite{WangIjcai19, LiSIGIR23} focus on leveraging Graph Neural Networks (GNNs) \cite{GNN} to model intent transitions and spatio-temporal features. \citet{LiKDD22} introduce AutoIntent, featuring disentangled intent encoders and intent discovery decoders. They construct dual hyper-graphs to capture relationships and intent features. \citet{ping21kdd} propose an intent detection and prediction system combining human expert knowledge and consumption information to capture user preferences and context. With the rise of large language models (LLM), researchers have begun to use LLM agents to simulate behavioral intents \cite{gao2023large, yuan2024unist}. \citet{shao2024beyond} develop an LLM workflow named Chain-of-Planned Behaviour for mobility behavior generation, which reflects the important spatial-temporal dynamics of human activities.
To solve the problem of insufficient user data, Yuan et al. \cite{yuan2024generating,yuan2023learning} motivated Maslow's need theory, propose a knowledge-driven simulation framework based on generative adversarial imitation learning. 

However, the above methods only cover some scenes in daily life resulting in the user behaviors being discontinuous and incomplete. Therefore, they can not deeply explore the user's common patterns and individual differences behind the user behavior sequences.

\subsection{On-device Recommendation Model}
Device-side recommender systems diverge from cloud-side recommendations by transferring model processing from the cloud to the device. This paradigm encompasses three primary approaches: (1) Device-side deployment, where models are trained in the cloud and deployed directly onto devices \cite{WangIjcai19}. (2) Device-side learning, where models are trained directly on devices, often employing collaborative learning \cite{GuoMWUT21}. (3) Device-cloud collaboration, integrating devices with cloud-based models to enhance performance \cite{yao2021device,lv2023duet,wang2023cloud,zheng2024decentralized}. \citet{yan2022devicebase} propose MPDA, which augments the user's local data by retrieving similar data from the cloud's pool. \citet{ding2023dc} introduce a collaborative learning framework that vertically divides the base model into two submodels: a larger one for cloud-side samples and a smaller one for device-side data, incorporating the output of the larger model.
 
 Recent studies integrate meta-learning into recommendations to learn shared global meta-parameters to quickly adapt to individual user-specific parameters\cite{ pan22meta}.
We et al.~\cite{Wei22KDDCLOVER} propose CLOVER, a comprehensive fair meta-learning framework, which introduces a multi-task adversarial learning scheme to satisfy fairness. \citet{Kim23MetaBert4Rec} propose a recommendation framework based on gradient-based meta-learning that captures the imbalanced rating distribution of each user and computes adaptive loss for user-specific learning. 

However, the above methods do not consider the difference in the distribution of cloud data and device data, which is not conducive to personalized learning.

\subsection{Cross-domain Fine-tuning of Pretrained LM}

This year we have witnessed rapid advancements in NLP foundation models, with increasing applications of LLMs in recommendation. Two main paradigms emerge: (1) Prompt tuning, where contextual tokens guide the model's response \cite{lester2021prompttuning}. \citet{Geng22P5} propose P5 first employ LLMs in a unified text-to-text approach. (2) Instruction tuning involves detailed text instructions to enhance zero-shot model performance \cite{brown2020instructtuning}. \citet{Bao23Tallrec} propose TALLRec, align LLMs with recommendations through data tuning, \citet{wei2023llmrec} present LLMRec, enhances systems via LLM-based graph augmentation.

Moreover, the transformer, a fundamental component of LLM, tokenizes inputs into embeddings, endowing it with universal representation for cross-domain transfer. \citet{LuAAAI22FLM} illustrates that PLM enhances performance and computational efficiency in non-language downstream tasks. \citet{zhou2023onefitsall} offers a unified framework for diverse time series tasks, showing that PLM yields comparable performance across main time series analysis tasks. \citet{jin2023timellm} introduce Time-LLM, a reprogramming framework for general time series forecasting, aligning time series with text prototypes to reconcile two modalities. \citet{liu2023unitime} propose UniTime for multivariate time series forecasting, employing domain instructions and a language-TS transformer to achieve zero-shot transferability through modality alignment.

LLMs harness a rich dataset of human behaviors during training, encompassing prevalent patterns, common sense, and underlying rules, yet the application of LLMs in simulating human behavior and user intent prediction remains an unexplored territory.
\section{conclusion}
Our research adapting PLMs into the human behavioral domain for on-device user intent prediction. We propose a population-to-individual tuning framework, which contains two main stages. In the population-level tuning stage, we leverage a PLM to capture the population-level common behavior patterns with the event reconstruction loss to enhance the event-to-intent transition pattern and obtain a lightweight predictor by model distillation. In the individual-level tuning framework, we utilize adaptive unlearning to correct the bias in long-tail intents due to the inconsistency between the intent distribution on population-level and individual-level. Finally, we use the individual user data to finetune and derive a personalized intent prediction model. 

In future work, we aim to extend the number of intents and use disentanglement methods \cite{quan2023alleviating} to implement debiased learning to solve the problem of insufficient learning of long-tail intents. Besides, we aim to consider the semantics to enhance behavior understanding and prediction by urban knowledge graph \cite{liu2023urban,liu2023urbankg}.

\section*{Acknowledgments}
This research has been supported in part by BNRist, National Key Research and Development Program of China under Grant 2022YFB3104702; in part by the National Natural Science Foundation of China under Grant 62272262 and Grant U23B2030; in part by the joint project of Honor Inc. \& Tsinghua University.

\bibliographystyle{ACM-Reference-Format}
\balance
\bibliography{Reference}

\appendix
\section{Details of Modle Distillation}
\label{distilling}
In the distillation process, the teacher model produces soft targets, essentially probability distributions across intents. The student model is then trained to mimic these soft targets, rather than the actual outputs of the teacher model. This allows the student model to learn from the teacher model's knowledge without needing to replicate the same level of computational complexity \cite{hinton2014distilling}. Figure \ref{fig:distilling} shows the specific process of the model distilling.

\begin{figure}[h]
\vspace{-3mm}
    \centering
    \includegraphics[width = 0.92\linewidth]{ 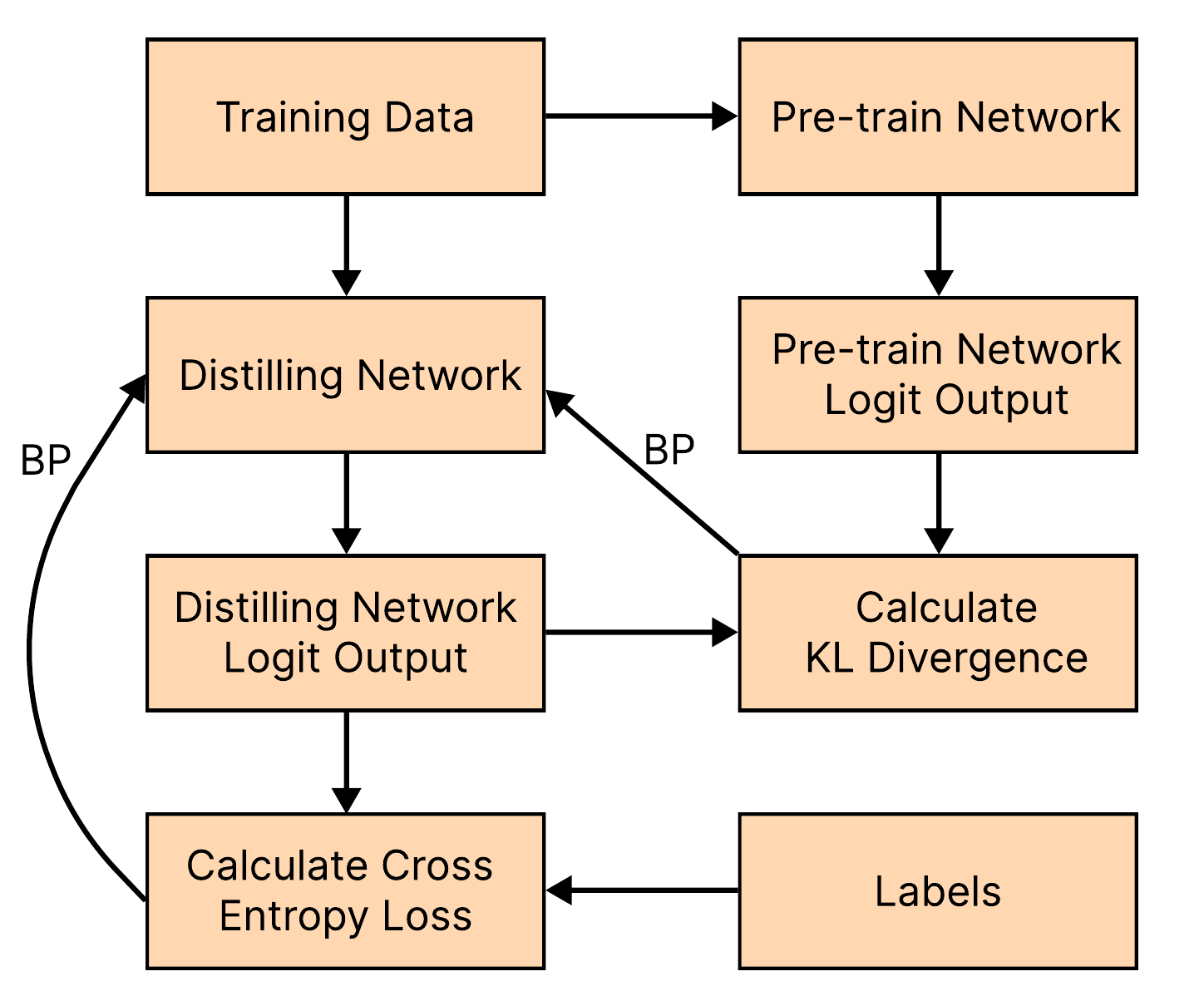}
    \vspace{-3mm}
    \caption{The process of model distilling.}
    \label{fig:distilling}
    \end{figure} 

\section{implementation details for reproducibility.}
\label{hp}
Here we provide detailed values of the hyperparameters in Table \ref{tab:hp}for reproducibility.

\begin{table}[h]
\centering
\vspace{-2mm}
\small
\caption{ Values of the hyperparameters}
{
\begin{tabular}{c  c c  } 
 \toprule
 &Hyperparameters & Value  \\
\midrule
 \multirow{5}*{\makecell[c]{Population-level \\tuning}}&Pretrain LM & GPT2-small \\
 &layer of transformer block & 12 \\
 &dimension of transformer block & 768\\
& $\alpha$ & 0.5 \\
& $\epsilon$ & 1 \\
\midrule
  \multirow{3}*{\makecell[c]{Individual-level \\tuning}}&layer of transformer block & 4 \\
 &dimension of transformer block & 768\\
 & $\lambda$ & 1 \\ 
  \bottomrule
 
\end{tabular}}
\label{tab:hp}
\end{table}

\section{Details of Baselines.}
\label{baseline}
Here we introduce the details of each baseline.
\begin{packed_itemize}
\item \textbf{CLOVER~\cite{Wei22KDDCLOVER}.}
CLOVER is a meta-learned recommendation models which consider three kinds of fairness: individual fairness, counterfactual fairness and group fairness through an adversarial learning method.

\item \textbf{MetaBert4Rec~\cite{Kim23MetaBert4Rec}.}
MetaBert4Rec is a sequential recommendation framework based on gradient-based meta-learning to capture the imbalanced rating distribution of each user.

\item \textbf{P5~\cite{Geng22P5}.}
P5 is the first work to propose a unified paradigm that integrates various recommendation-related tasks into a shared conditional language generation framework.

\item \textbf{InstructRec~\cite{Zhang2023Insprom}.}
InstrucRec considers that preferences or needs can be expressed in natural language descriptions so that LLM can understand and execute the instruction for fulfilling the recommendation task.

\item \textbf{LSAT~\cite{Shi2023LSAT}.}
LSAT utilize two adaptation LoRA modules to learn long-term and short-term user preferences separately and then integrates them to merge the different types of preferences.

\item \textbf{One fits All (OFA)~\cite{zhou2023onefitsall}}
OFA is a unified framework that uses a frozen pre-train language model to investigate cross-modality knowledge transfer for time series forecasting tasks.

\item \textbf{TallRec~\cite{Bao23Tallrec}}
TallRec is an efficient Tuning framework for Aligning LLMs with Recommendations, which structures the recommendation data as instructions and tunes the LLM via an additional instruction tuning process.

\item \textbf{EODRec~\cite{Xia23DeviceSession}}
EODRec is an ultra-compact efficient on-device session-based recommendation that integrates discrete compositional code learning into recommendation systems to compress an item embedding table.

\item \textbf{MPDA~\cite{yan2022devicebase}}
MPDA is a new device-cloud collaborative learning framework whose general idea is to retrieve some similar data from the cloud's global pool to augment the user's local data as the target domain. We choose the OFA model as the backbone.

\end{packed_itemize}

\section{Details of Metrics.}
\label{metrics}
we employ five widely used metrics: weighted precision ($Prec_w$), weighted recall ($Rec_w$), macro precision ($Prec_m$), macro recall ($Rec_m$), and NDCG(N). The calculation of each metric is as follows.
The formula for $Prec_w$ :
\begin{equation}\label{equ:Prec_w}
    Prec_w = \frac{\sum_{c \in C} (\text{TP}_c + \text{FP}_c) \cdot \text{Precision}_c}{\sum_{c \in C} (\text{TP}_c + \text{FP}_c)}
\end{equation}
The formula for $Rec_w$ :
\begin{equation}\label{equ:Rec_w}
    Rec_w = \frac{\sum_{c \in C} (\text{TP}_c + \text{FN}_c) \cdot \text{Recall}_c}{\sum_{c \in C} (\text{TP}_c + \text{FN}_c)}
\end{equation}
The formula for $Prec_m$ :
\begin{equation}\label{equ:Prec_m}
    Prec_m = = \frac{1}{|C|} \sum_{c \in C} \frac{\text{TP}_c}{\text{TP}_c + \text{FP}_c}
\end{equation}
The formula for $Rec_m$ :
\begin{equation}\label{equ:Rec_m}
    Rec_m = \frac{1}{|C|} \sum_{c \in C} \frac{\text{TP}_c}{\text{TP}_c + \text{FN}_c}
\end{equation}
Where $|C|$ represents the total number of classes, True Positives $\text({TP}_c)$ denotes the number of samples correctly classified as class $c$, False Positives $\text({FP}_c)$ represents the number of samples incorrectly classified as class $c$, and False Negatives $\text({FN}_c)$ stands for the number of samples incorrectly classified as other classes instead of class $c$. And $\text{Precision}_c$ and $\text{Recall}_c$ respectively refer to the precision and recall of class $c$.\\
The formula for $N@k$ :
\begin{equation}\label{equ:N@K}
    N@k = \frac{\sum_{i=1}^{K} \frac{2^{rel_i} - 1}{\log_2(i+1)}}{\sum_{j=1}^{|REL_K|} \frac{rel_j - 1}{\log_2(j+1)}}
\end{equation}
where $rel_i$ means the graded relevance of the result at position $i$, and $|REL_K|$ means the list of predictions in the result ranking list up to position $K$.

\section{Details of the Tree Model}
\label{treemodel}
We employ LightGBM \cite{ke2017lightgbm} as our tree model, a method widely used in competitions. The hyperparameters we use for the model are shown in the table below.

\begin{table}[h]
\centering
\vspace{-2mm}
\small
\caption{ Hyperparameters Setting of LightGBM  }
{
\begin{tabular}{c  c c  } 
 \toprule
 \multicolumn{2}{c}{Hyperparameters} & Value  \\
 \midrule

 \multicolumn{2}{c}{objective} & multiclass  \\
 \multicolumn{2}{c}{boosting} & gbdt  \\
 \multicolumn{2}{c}{num class} & 18 \\
 \multicolumn{2}{c}{num iterations} & 2000 \\
 \multicolumn{2}{c}{num leaves} & 32 \\
 \multicolumn{2}{c}{max depth} & -1 \\
 \multicolumn{2}{c}{min data in leaf} & 20 \\
 \multicolumn{2}{c}{feature fraction} & 1 \\
 \multicolumn{2}{c}{early stopping round} & 75 \\
 \multicolumn{2}{c}{$\lambda\_l1$} & 0 \\
 \multicolumn{2}{c}{$\lambda\_l2$} & 0 \\
 \multicolumn{2}{c}{random state} & 42 \\
 
  \bottomrule
 
\end{tabular}}
\label{tab:treeFeature}
\end{table}

Please note that if the size of the dataset is less than 200, we will reduce the complexity of LightGBM by setting max depth=3, num leaves=3, $\lambda\_l1$=1, $\lambda\_l2$=1.

Our feature Set is shown below:
\begin{enumerate}
    \item Output probabilities of all categories from GPT-2.
    \item Position features (whether in the top 10 frequent locations).
    \item Current hour, current day of the week, current timestamp, and indicators for morning/afternoon/evening and weekday/weekend.
    \item For each category (illustrated by event $e$), time difference $T_0$ between the current time and the time of the last occurrence of event $e$, time difference $T_1$ between the time of the last occurrence of event $e$ and the time of the second-to-last occurrence of event $e$.
Given $x_i=(u_i,l_i,t_i,e_i)$, suppose event $e$ occurred $n$ times before the current event $t_i$. The time of the $n_{th}$ occurrence of event $e$ is denoted as $t_{en}$.
The explanation of the mathematical formula for the time difference is as follows
\begin{equation}\label{equ:treeFeature1}
    T_0 = t_i - t_{en}
\end{equation}
\begin{equation}\label{equ:treeFeature2}
    T_1 = t_{en} - t_{e(n-1)}
\end{equation}
\end{enumerate}



\end{document}